\def\year{2022}\relax
\documentclass[letterpaper]{article} %
\usepackage{aaai22}  %
\usepackage{times}  %
\usepackage{helvet}  %
\usepackage{courier}  %
\usepackage[hyphens]{url}  %
\usepackage{graphicx} %
\usepackage{natbib}  %
\usepackage{caption} %
\DeclareCaptionStyle{ruled}{labelfont=normalfont,labelsep=colon,strut=off} %
\usepackage{algorithm}
\usepackage[noend]{algpseudocode}
\usepackage{listings}

\usepackage{newfloat}
\usepackage{listings}
\floatstyle{ruled}
\newfloat{listing}{tb}{lst}{}
\floatname{listing}{Listing}
\usepackage{booktabs}
\usepackage{xcolor}
\usepackage{amsthm, amsmath, amsfonts, amssymb}
\usepackage{pifont}

\newcommand{\yes}{\ding{51}}
\newcommand{\no}{\ding{55}}

\definecolor{teal}{HTML}{00897B}

\newcommand\R{\mathbb{R}}
\newcommand\E{\mathbb{E}}
\newcommand\N{\mathbb{N}}
\newcommand\fmsg{\overrightarrow m}
\newcommand\bmsg{\overleftarrow m}
\newcommand\Fmsg{\overrightarrow M}
\newcommand\Bmsg{\overleftarrow M}

\title{Introducing Symmetries to Black Box Meta Reinforcement Learning}
\author{
Louis Kirsch\textsuperscript{\rm 1 \rm 2},
Sebastian Flennerhag\textsuperscript{\rm 1},
Hado van Hasselt\textsuperscript{\rm 1},
Abram Friesen\textsuperscript{\rm 1},
Junhyuk Oh\textsuperscript{\rm 1},\\
Yutian Chen\textsuperscript{\rm 1}
}
\affiliations{
    \textsuperscript{\rm 1} DeepMind\\
    \textsuperscript{\rm 2} The Swiss AI Lab IDSIA, USI, SUPSI\\
    louis@idsia.ch, \{flennerhag,hado,abef,junhyuk,yutianc\}@deepmind.com
}

\begin{document}

\maketitle

\begin{abstract}

Meta reinforcement learning (RL) attempts to discover new RL algorithms automatically from environment interaction.
In so-called black-box approaches, the policy and the learning algorithm are jointly represented by a single neural network.
These methods are very flexible, but they tend to underperform compared to human-engineered RL algorithms in terms of generalisation to new, unseen environments.
In this paper, we explore the role of symmetries in meta-generalisation.
We show that a recent successful meta RL approach that meta-learns an objective for backpropagation-based learning exhibits certain symmetries (specifically the reuse of the learning rule, and invariance to input and output permutations) that are not present in typical black-box meta RL systems.
We hypothesise that these symmetries can play an important role in meta-generalisation.
Building off recent work in black-box supervised meta learning, we develop a black-box meta RL system that exhibits these same symmetries.
We show through careful experimentation that incorporating these symmetries can lead to algorithms with a greater ability to generalise to unseen action \& observation spaces, tasks, and environments.

\end{abstract}

\section{Introduction}

Recent work in meta reinforcement learning (RL) has begun to tackle the challenging problem of automatically discovering general-purpose RL  algorithms~\citep{kirsch2019improving,alet2020meta,oh2020discovering}. 
These methods learn to reinforcement learn by optimizing for earned reward over the lifetimes of many agents in multiple environments.
If the discovered learning principles are sufficiently general-purpose, then the learned algorithms should generalise to significantly different unseen environments. 
Depending on the structure of the learned algorithm, these methods can be partitioned into  backpropagation-based methods, which learn to use the backpropagation algorithm to reinforcement learn, and black-box-based methods, in which a single (typically recurrent) neural network jointly specifies
the agent and RL algorithm~\citep{wang2016learning,duan2016rl}.
While backpropagation-based methods are more prevalent due to their relative ease of implementation and theoretical guarantees, black-box methods are expressive and have the potential to avoid some of the issues with backpropagation-based optimization, such as memory requirements, catastrophic forgetting, and differentiability. %

Unfortunately, black-box methods have not yet been successful at discovering general-purpose RL algorithms that compete with the generality of human-engineered algorithms.
In this work, we show that black-box methods exploit fewer symmetries than backpropagation-based methods.
We hypothesise that introducing more symmetries to black-box meta-learners can improve their generalisation capabilities.
We test this hypothesis by introducing a number of symmetries into an existing black-box meta learning algorithm, including
(1) the use of the same learned learning rule across all nodes of the neural network (NN),
(2) the flexibility to work with any input, output, and architecture sizes,
and (3) invariance to permutations of the inputs and outputs (for dense layers).
Permutation invariance implies that for any permutation of inputs and outputs the learning algorithm produces the same policy.
As we show, this is similar to dense NNs trained with backpropagation that also exhibit permutation invariance.
We refer to such agents as \emph{symmetric learning agents} (SymLA).

To introduce these symmetries, we build on variable shared meta learning (VSML)~\citep{kirsch2020meta}, which we adapt to the RL setting.
VSML arranges multiple RNNs like weights in a NN and performs message passing between these RNNs.
We then perform meta training and meta testing similar to black-box MetaRNNs, also known as RL$^2$~\citep{wang2016learning,duan2016rl}.
We experimentally validate SymLA on bandits, classic control, and grid worlds, comparing generalisation capabilities to MetaRNNs.
SymLA improves generalisation when varying action dimensions, permuting observations and actions, and significantly changing tasks and environments.

\begin{figure*}
    \centering
    \includegraphics[width=0.92\textwidth]{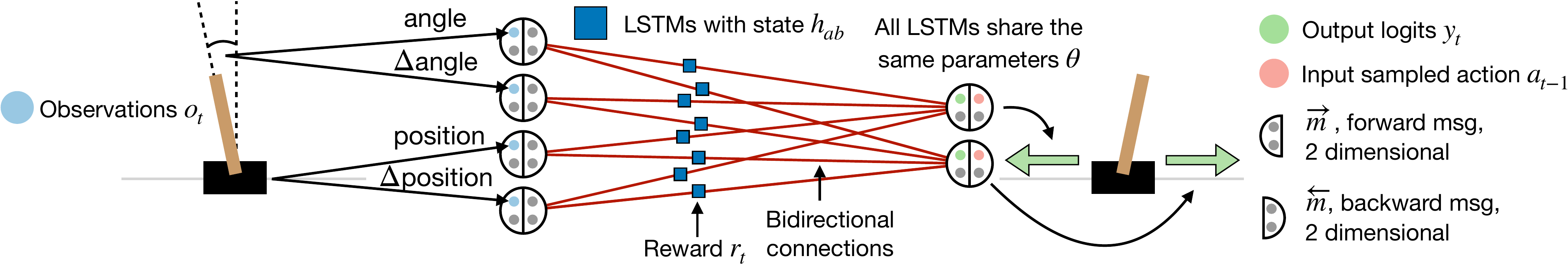}
    \caption{
    The architecture for the proposed \emph{symmetric learning agents} (SymLA) that we use to investigate black-box learning algorithm with symmetries.
    Weights in a neural network are replaced with small parameter-shared RNNs.
    Activations in the original network correspond to messages passed between RNNs, both in the forward $\protect\fmsg$ and backward $\protect\bmsg$ direction in the network.
    These messages may contain external information such as the environment observation, previously taken actions, and rewards from the environment.
    }
    \label{fig:architecture}
\end{figure*}

\section{Preliminaries}\label{sec:background}

\subsection{Reinforcement Learning}
The RL setting in this work follows the standard (PO)MDP formulation.
At every time step, $t = 1,2,\ldots$ the agent receives a new observation $o_t \in \mathcal{O}$ generated from the environment state $s_t \in \mathcal{S}$ and performs an action $a_t \in \mathcal{A}$ sampled from its (recurrent) policy  $\pi_\theta = p(a_t|o_{1:t},a_{1:t-1})$.
The agent receives a reward $r_t \in \mathcal{R} \subset \mathbb{R}$ and the environment transitions to the next state.
This transition is defined by the environment dynamics $e = p(s_{t+1}, r_t|s_t, a_t)$.
The initial environment state $s_1$ is sampled from the initial state distribution $p(s_1)$.
The goal is to find the optimal policy parameters $\theta^*$ that maximise the expected return $R = \mathbb{E}[\sum_{t=1}^{T} \gamma^t r_t]$ where $T$ is the episode length, and $0 < \gamma \leq 1$ is a discount factor ($T = \infty$, $\gamma < 1$ for non-episodic MDPs).

\subsection{Meta Reinforcement Learning}\label{sec:mrl}

The meta reinforcement learning setting is concerned with discovering novel agents that learn throughout their multi-episode lifetime ($L \geq T$) by making use of rewards $r_t$ to update their behavior.
This can be formulated as maximizing $\mathbb{E}_{e \sim p(e)}[\mathbb{E}[\sum_{t=1}^{L} \gamma^t r_t]]$ where $p(e)$ is a distribution of meta-training environments.
The objective itself is similar to a multi-task setting.
In this work, we discuss how the structure of the agent influences the degree to which it \emph{learns} and \emph{generalises} in novel tasks and environments.
We seek to discover \emph{general-purpose} learning algorithms that generalise outside the meta-training distribution.

We can think of an agent that learns throughout its lifetime as a history-dependent map $a_t, h_t = f(h_{t-1}, o_t, r_{t-1}, a_{t-1})$ that produces an action $a_t$ and new agent state $h_t$ given its previous state $h_{t-1}$, an observation $o_t$, environment reward $r_{t-1}$, and previous action $a_{t-1}$.
In the case of backpropagation-based learning, $f$ is decomposed into:
(1) a \emph{stationary} policy $\pi_\theta^{(s)}$ that maps the current state into an action, $a_t = \pi_\theta^{(s)}(o_t)$; and
(2) a backpropagation-based update rule that optimizes a given objective $J$ by propagating the error signal backwards and updating the policy in fixed intervals (e.g. after each episode).
In its simplest form, for any dense layer $k \in \{1, \ldots, K\}$ of a NN policy with size $A^{(k)} \times B^{(k)}$, inputs $x^{(k)}$, outputs $x^{(k+1)}$, and weights $w^{(k)} \subset \theta$, the backpropagation update rule is given by
\begin{align}
    x^{(k+1)}_b &= \sum_a x^{(k)}_a w^{(k)}_{ab} & \textrm{(forward pass)} \label{eq:fwd_backprop} \\
    \delta^{(k-1)}_a &= \sum_b \delta_b^{(k)} w_{ab}^{(k)} & \textrm{(backward pass)} \label{eq:bwd_backprop} \\
    \Delta w_{ab}^{(k)} &= -\alpha\frac{\partial J}{\partial w_{ab}^{(k)}} = -\alpha x_a^{(k)} \delta_b^{(k)} & \textrm{(update)}\label{eq:meta-rl-backprop}
\end{align}
where $a \in \{1, \ldots, A^{(k)}\}$, $b \in \{1, \ldots, B^{(k)}\}$, $\alpha$ is the learning rate, $\delta$ are error terms, and the agent state $h$ corresponds to parameters $\theta$.
The initial error is given by the gradient at the NN outputs, $\delta^{(k)} = \frac{\partial J}{\partial x^{(K+1)}}$.
Transformations such as non-linearities are omitted here.
Works in meta-reinforcement learning that take this approach parameterise the objective $J_\phi$ and meta-learn its parameters \citep{kirsch2019improving,oh2020discovering}.

In contrast, black-box meta RL~\citep{duan2016rl,wang2016learning} meta-learns $f$ directly in the form of a single \emph{non-stationary} policy $\pi_\theta$ with memory.
Parameters of $f$ represent the learning algorithm (no explicit $J_\phi$) while the state $h$ represents the policy.
In the simplest form of an RNN representation of $f$, given a current hidden state $h$ and inputs $o,r,a$ (concatenated $[\cdot]$), updates to the policy take the form
\begin{equation}\label{eq:meta-rl-blackbox}
a_b, h_b \leftarrow f_{\theta}(h, o, r, a)_{b} = \sigma(\sum_a [h, o, r, a]_a v_{ab}),
\end{equation}
with parameters $\theta = v$ and activation function $\sigma$, omitting the bias term.
We refer to this as the MetaRNN.
The inputs must include, beyond the observation $o$, the previous reward $r$ and action $a$, so that the meta-learner can learn to associate past actions with rewards~\citep{schmidhuber1993self,wang2016learning}.
Further, black-box systems do not reset the state $h$ between episode boundaries, so that the learning algorithm can accumulate knowledge through the agent's lifetime.

\section{Symmetries in Meta RL}

In this section, we demonstrate how the learning dynamics in backpropagation-based systems (Equation \ref{eq:meta-rl-backprop}) differ from the learning dynamics in black-box systems (Equation \ref{eq:meta-rl-blackbox}), and how this affects the generalisation of black-box methods to novel environments.

\subsection{Symmetries in Backpropagation-based Meta RL}\label{sec:symmetries_backprop}

We first identify symmetries that backpropagation-based systems exhibit and discuss how they affect the generalisability of the learned learning algorithms.
\begin{enumerate}
    \item \textbf{Symmetric learning rule.}
    In Equation \ref{eq:meta-rl-backprop}, each parameter $w_{ab}$ is updated by the same update rule based on information from the forward and backward pass.
    Meta-learning an objective $J_\phi$ affects the updates of each parameter symmetrically through backpropagation.
    \item \textbf{Flexible input, output, and architecture sizes.}
    Because the same rule is applied everywhere, the learning algorithm can be applied to arbitrarily sized neural networks, including variations in input and output sizes.
    This involves varying $A$ and $B$ and the number of layers, affecting how often the learning rule is applied and how many parameters are being learned.
    \item \textbf{Invariance to input and output permutations.}
    Given a permutation of inputs and outputs in a layer, defined by the bijections $\rho: \N \to \N$ and $\rho': \N \to \N$, the learning rule is applied as
    $x^{(k+1)}_{\rho'(b)} = \sum_a x^{(k)}_{\rho(a)} w^{(k)}_{ab}$,
    $\delta^{(k-1)}_{\rho(a)} = \sum_b \delta^{(k)}_{\rho'(b)} w^{(k)}_{ab}$, and
    $\Delta w^{(k)}_{ab} = -\alpha x^{(k)}_{\rho(a)} \delta^{(k)}_{\rho'(b)}$.
    Let $w'$ be a weight matrix with $w'^{(k)}_{\rho(a)\rho'(b)} = w^{(k)}_{a,b}$, then we can equivalently write
    $x^{(k+1)}_{\rho'(b)} = \sum_a x^{(k)}_{\rho(a)} w'^{(k)}_{\rho(a)\rho'(b)}$,
    $\delta^{(k-1)}_{\rho(a)} = \sum_b \delta^{(k)}_{\rho'(b)} w'^{(k)}_{\rho(a)\rho'(b)}$, and
    $\Delta w'^{(k)}_{\rho(a)\rho'(b)} = -\alpha x^{(k)}_{\rho(a)} \delta^{(k)}_{\rho'(b)}$.
    If all elements of $w'^{(k)}$ are initialized i.i.d., we can interchangeably use $w$ in place of $w'$ in the above updates.
    By doing so, we recover the original learning rule equations for any $a, b$.
    Thus, the learning algorithm is invariant to input and output permutations.
\end{enumerate}
While backpropagation has inherent symmetries, these symmetries would be violated if the objective function $J_\phi$ would be asymmetric.
Formally, when permuting the NN outputs $y = x^{(K+1)}$ such that $y'_b = y_{\rho'(b)}$, $J_\phi$ should satisfy that the gradient under the permutation is also a permutation
\begin{equation}
    \frac{\partial J_\phi(y')}{\partial y'_b} = \left[ \frac{\partial J_\phi(y)}{\partial y} \right]_{\rho'(b)}
\end{equation}
where the environment accepts the action permuted by $\rho'$ in the case of $J_\phi(y')$.
This is the case for policy gradients, for instance, if the action selection $\pi(a|s)$ is permuted according to $\rho'$.
When meta-learning objective functions, prior work carefully designed the objective function $J_\phi$ to be symmetric.
In MetaGenRL~\citep{kirsch2019improving}, taken actions were processed element-wise with the policy outputs and sum-reduced by the loss function.
In LPG~\citep{oh2020discovering}, taken actions and policy outputs were not directly fed to $J_\phi$, but instead only the log probability of the action distribution was used.

\subsection{Insufficient Symmetries in Black-box Meta RL}

Black-box meta learning methods are appealing as they require few hard-coded biases and are flexible enough to represent a wide range of possible learning algorithms.
We hypothesize that this comes at the cost of the tendency to overfit to the given meta training environment(s) resulting in overly specialized learning algorithms.

Learning dynamics in backpropagation-based systems (Equation \ref{eq:meta-rl-backprop}) differ significantly from learning dynamics in black-box systems (Equation \ref{eq:meta-rl-blackbox}).
In particular, meta-learning $J_\phi$ is significantly more constrained, since $J_\phi$ can only indirectly affect each policy parameter $w_{ab}^{(k)}$ through the \emph{same} learning rule from Equation \ref{eq:meta-rl-backprop}.
In contrast, in black-box systems (Equation \ref{eq:meta-rl-blackbox}), each policy state $h_b$ is directly controlled by \emph{unique} meta-parameters (vector $v_{\cdot b}$), thereby encouraging the black-box meta-learner to construct specific update rules for each element of the policy state.
This results in sensitivity to permutations in inputs and outputs.
Furthermore, input and output spaces must retain the same size as those are directly dependent on the number of RNN parameters.

As an example, consider a meta-training distribution of two-armed bandits where the expected payout of the first arm is much larger than the second.
If we meta-train a MetaRNN on these environments then
when meta-testing the MetaRNN will have learned to immediately
increase the probability of pulling the first arm, independent of any observed rewards.
If instead the action probability is adapted using REINFORCE or a meta-learned symmetric objective function then, due to the implicit symmetries, the learning algorithm could not differentiate between the two arms to favor one over the other.
While the MetaRNN behavior is optimal when meta-testing on the same meta-training distribution, it completely fails to generalise to other distributions.
Thus, the MetaRNN results in a non-learning, biased solution, whereas the backpropagation-based approach results in a learning solution.
In the former case, the learning algorithm is overfitted to only produce a fixed policy that always samples the first arm.
In the latter case, the learning algorithm is unbiased and will learn a policy from observed rewards to sample the first arm.
Beyond bandits, for reasonably sized meta-training distributions, we may have any number of biases in the data that a MetaRNN will inherit, impeding generalisation to unseen tasks and environments.

\section{Adding Symmetries to Black-box Meta RL}

A solution to the illustrated over-fitting problem with black-box methods is the introduction of symmetries into the parameterisation of the policy.
This can be achieved by generalising the forward pass (Equation \ref{eq:fwd_backprop}), backward pass (Equation \ref{eq:bwd_backprop}), and element-wise update (Equation \ref{eq:meta-rl-backprop}) to parameterized versions.
We further subsume the loss computation into these parameterized update rules.
Together, they form a single recurrent policy with additional symmetries.
Prior work on variable shared meta learning (VSML)~\citep{kirsch2020meta} used similar principles to meta-learn supervised learning algorithms.
In the following, we extend their approach to deal with the RL setting.

\subsection{Variable Shared Meta Learning}

VSML describes neural architectures for meta learning with parameter sharing.
This can be motivated by meta learning how to update weights~\citep{bengio1992optimization,schmidhuber1993reducing} where the update rule is shared across the network.
Instead of designing a meta network that defines the weight updates explicitly, we arrange small parameter-shared RNNs (LSTMs) like weights in a NN and perform message passing between those.

In VSML, each weight $w_{ab}$ with $w \in \mathbb{R}^{A \times B}$ in a NN is replaced by a small RNN with parameters $\theta$ and hidden state $h_{ab} \in \R^{N}$.
We restrict ourselves to dense NN layers here, where $w$ corresponds to the weights of that layer with input size $A$ and output size $B$.
This can be adapted to other architectures such as CNNs if necessary.
All these RNNs share the same parameters $\theta$, defining both what information propagates in the neural network, as well as how states are updated to implement learning.
Each RNN with state $h_{ab}$ receives the analogue to the previous activation, here called the vectorized forward message $\fmsg_a \in \R^{\Fmsg}$, and the backward message $\bmsg_b \in \R^{\Bmsg}$ for information flowing backwards in the network (asynchronously).
The backward message may contain information relevant to credit assignment, but is not constrained to this.
The RNN update equation (compare Equation \ref{eq:meta-rl-backprop} and \ref{eq:meta-rl-blackbox}) is then given by
\begin{equation}\label{eq:vsml_state_update}
    h_{ab}^{(k)} \leftarrow f_{\textrm{RNN}}(h_{ab}^{(k)}, \fmsg_a^{(k)}, \bmsg_b^{(k)})
\end{equation}
for layer $k$ where $k \in \{1, \ldots, K\}$ and $a \in \{1, \ldots, A^{(k)}\}, b \in \{1, \ldots, B^{(k)}\}$.
Similarly, new forward messages are created by transforming the RNN states using a function $f_{\fmsg}: \R^N \to \R^{\Fmsg}$ (compare Equation \ref{eq:fwd_backprop}) such that
\begin{equation}\label{eq:gen_fwd_msg}
    \fmsg_b^{(k+1)} = \sum_a f_{\fmsg}(h_{ab}^{(k)})
\end{equation}
defines the new forward message for layer $k + 1$ with $b \in \{1, \ldots, B^{(k)} = A^{(k+1)}\}$.
The backward message is given by $f_{\bmsg}: \R^N \to \R^{\Bmsg}$ (compare Equation \ref{eq:bwd_backprop}) such that
\begin{equation}\label{eq:gen_bwd_msg}
    \bmsg_a^{(k-1)} = \sum_b f_{\bmsg}(h_{ab}^{(k)})
\end{equation}
and $a \in \{1, \ldots, A^{(k)} = B^{(k-1)}\}$.
For simplicity, we use $\theta$ below to denote all of the VSML parameters, including those of the RNN and forward and backward message functions.

In the following, we derive a black-box meta reinforcement learner based on VSML (visualized in Figure \ref{fig:architecture}).

\subsection{RL Agent Inputs and Outputs}
At each time step in the environment, the agent's inputs consist of the previously taken action $a_{t-1}$, current observation $o_t$ and previous reward $r_{t-1}$.
We feed $r_{t-1}$ as an additional input to each RNN, the observation $o_t \in \R^{A^{(1)}}$ to the first layer ($\fmsg_{\cdot1}^{(1)} := o_{t}$), and the action $a_{t-1} \in \{0,1\}^{B^{(K)}}$ (one-hot encoded) to the last layer ($\bmsg_{\cdot1}^{(K)} := a_{t-1}$).
The index $1$ refers to the first dimension of the $\Fmsg$ or $\Bmsg$-dimensional message.
We interpret the agent's output message $y = \fmsg_{\cdot1}^{(K+1)}$ as the unnormalized logits of a categorical distribution over actions.
While we focus on discrete actions only in our present experiments, this can be adapted for probabilistic or deterministic continuous control.

\subsection{Architecture Recurrence and Reward Signal}

Instead of using multiple layers ($K > 1$), in this paper we use a single layer ($K = 1$).
In Equation \ref{eq:vsml_state_update}, RNNs in the same layer can not coordinate directly as their messages are only passed to the next and previous layer.
To give that single layer sufficient expressivity for the RL setting, we make it `recurrent' by processing the layer's own messages $\fmsg_b^{(k+1)}$ and $\bmsg_a^{(k-1)}$.
The network thus has two levels of recurrence:
(1) Each RNN that corresponds to a weight of a standard NN and 
(2) messages that are generated according to Equation \ref{eq:gen_fwd_msg} and \ref{eq:gen_bwd_msg} and fed back into the same layer.
Furthermore, each RNN receives the current reward signal $r_{t-1}$ as input.
The update equation is given by
\begin{equation}\label{eq:state_update}
    h_{ab}^{(k)} \leftarrow f_{\textrm{RNN}}(h_{ab}^{(k)}, \underbrace{\fmsg_a^{(k)}, \bmsg_b^{(k)}, r_{t-1}}_{\textrm{environment inputs}}, \underbrace{\fmsg_b^{(k+1)}, \bmsg_a^{(k-1)}}_{\textrm{from previous step}})
\end{equation}
where $a \in \{1, \ldots, A^{(k)}\}, b \in \{1, \ldots, B^{(k)}\}$.
As we only use a single layer, $k = 1$,
we apply the update multiple times (multiple micro ticks) for each step in the environment. %
This can also be viewed as multiple layers with shared parameters, where parameters correspond to states $h$.
For pseudo code, see Algorithm \ref{alg:meta_training} in the appendix.

\begin{figure}
    \centering
    \includegraphics[width=0.9\columnwidth]{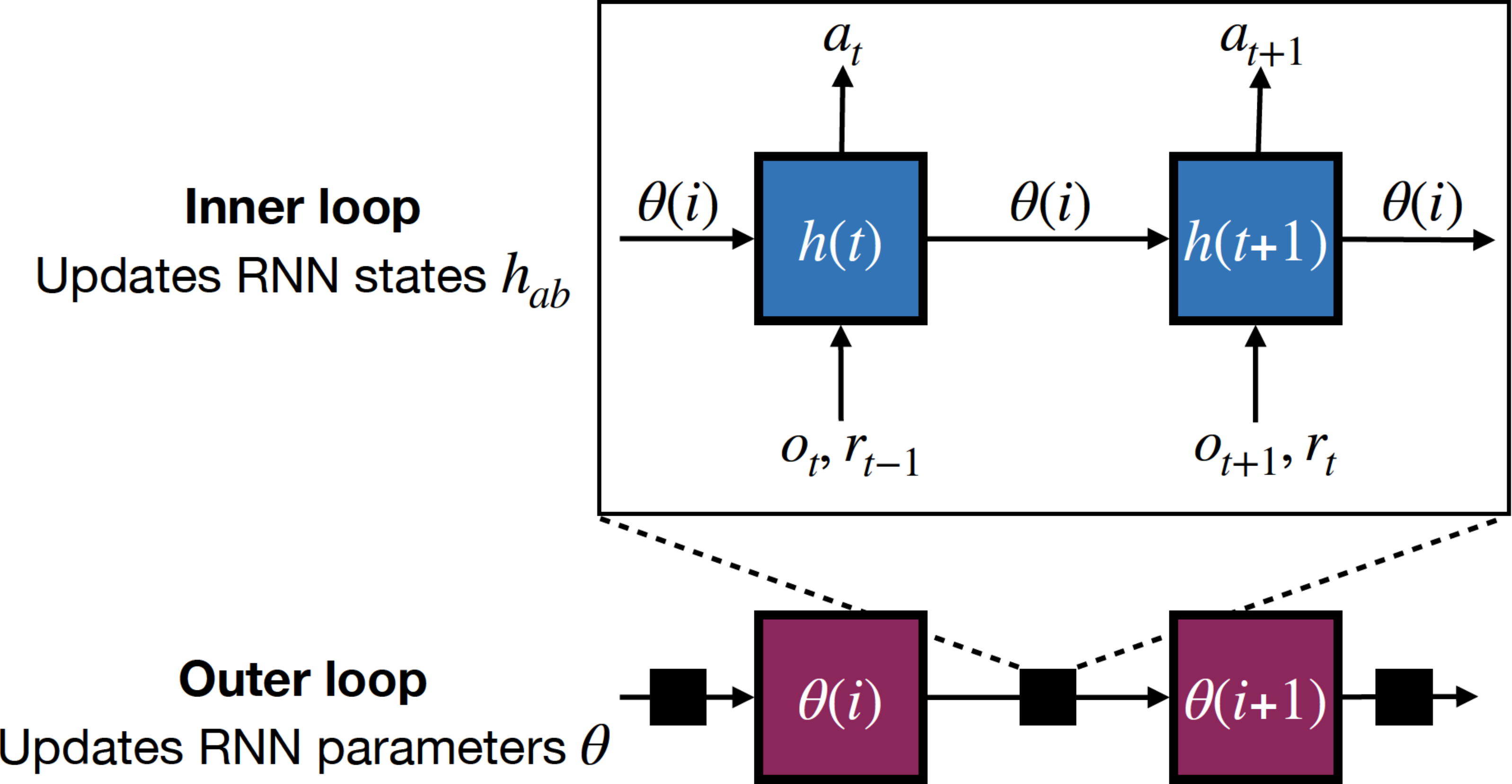}
    \caption{In SymLA, the inner loop recurrently updates all RNN states $h_{ab}(t)$ for agent steps $t \in \{1, \ldots, L\}$ starting with randomly initialized states $h_{ab}$.
    Based on feedback $r_t$, RNN states can be used as memory for learning.
    The learning algorithm encoded in the RNN parameters $\theta$ is updated in the outer loop by meta-training using ES.
    }
    \label{fig:inner-outer}
\end{figure}

\subsection{Symmetries in SymLA}

By incorporating the above changes to inputs, outputs, and architecture, we arrive at a black-box meta RL method with symmetries, here represented by our proposed \emph{symmetric learning agents} (SymLA).
By construction, SymLA exhibits the same symmetries as those described in Section \ref{sec:symmetries_backprop}, despite not using the backpropagation algorithm.
\begin{enumerate}
    \item \textbf{Symmetric learning rule.}
    The learning rule as defined by Equation \ref{eq:state_update} is replicated across $a \in \{1, \ldots, A\}$ and $b \in \{1, \ldots, B\}$ with the same parameter $\theta$.
    \item \textbf{Flexible input, output, and architecture sizes.}
    Changes in $A$, $B$, and $K$ correspond to input, output, and architecture size.
    This does not affect the number of meta-parameters and therefore these quantities can also be varied at meta-test time.
    \item \textbf{Invariance to input and output permutations.}
    When permuting messages using bijections $\rho$ and $\rho'$, the state update becomes
    $h_{ab}^{(k)} \leftarrow f_{\textrm{RNN}}(h_{ab}^{(k)}, \fmsg_{\rho(a)}^{(k)}, \bmsg_{\rho'(b)}^{(k)}, r_{t-1}, \fmsg_{\rho'(b)}^{(k+1)}, \bmsg_{\rho(a)}^{(k-1)})$, and the message transformations are $ \fmsg_{\rho'(b)}^{(k+1)} = \sum_a f_{\fmsg}(h_{ab}^{(k)})$ and $\bmsg_{\rho(a)}^{(k-1)} = \sum_b f_{\bmsg}(h_{ab}^{(k)})$.
    Similar to backpropagation, when RNN states $h_{ab}$ are initialized i.i.d., we can use $h_{\rho(a),\rho'(b)}$ in place of $h_{ab}$ to recover the original Equations \ref{eq:gen_fwd_msg}, \ref{eq:gen_bwd_msg}, \ref{eq:state_update}.
\end{enumerate}

\subsection{Learning / Inner Loop}
Learning corresponds to updating RNN states $h_{ab}$ (see Figure \ref{fig:inner-outer}).
This is the same as the MetaRNN~\citep{wang2016learning,duan2016rl} but with a more structured neural model.
For fixed RNN parameters $\theta$ which encode the learning algorithm, we randomly initialize all states $h_{ab}$.
Next, the agent steps through the environment, updating $h_{ab}$ in each step.
If the environment is episodic with $T$ steps, the agent is run for a lifetime of $L \geq T$ steps with environment resets in-between, carrying the agent state $h_{ab}$ over.

\subsection{Meta Learning / Outer Loop}
Each outer loop step unrolls the inner loop for $L$ environment steps to update $\theta$.
The SymLA objective is to maximize the agent's lifetime sum of rewards, i.e. $\sum_{t=1}^L r_t(\theta)$.
We optimize this objective using evolutionary strategies~\citep{wierstra2008natural,salimans2017evolution} by following the gradient
\begin{equation}\label{eq:evolution_strategies}
    \nabla_\theta \E_{\phi \sim \mathcal{N}(\phi|\theta, \Sigma)}[\E_{e \sim p(e)}[\sum_{t=1}^L r_t^{(e)}(\phi)]].
\end{equation}
with some fixed diagonal covariance matrix $\Sigma$ and environments $e \sim p(e)$.
We chose evolution strategies due to its ability to optimize over long inner-loop horizons without memory constraints that occur due to backpropagation-based meta optimization.
Furthermore, it was shown that meta-loss landscapes are difficult to navigate and the search distribution helps in smoothing those~\citep{metz2019understanding}.

\section{Experiments}

Equipped with a symmetric black-box learner, we now investigate how its learning properties differ from a standard MetaRNN.
Firstly, we learn to learn on bandits from \citet{wang2016learning} where the meta-training environments are similar to the meta-test environments.
Secondly, we demonstrate generalisation to unseen action spaces, applying the learned algorithm to bandits with varying numbers of arms at meta-test time---something that MetaRNNs are not capable of.
Thirdly, we demonstrate how symmetries improve generalisation to unseen observation spaces by creating permutations of observations and actions in classic control benchmarks.
Fourthly, we show how permutation invariance leads to generalisation to unseen tasks by learning about states and their associated rewards at meta-test time.
Finally, we demonstrate how symmetries result in better learning algorithms for unseen environments, generalising from a grid world to CartPole.
Hyper-parameters are in Appendix \ref{app:hyperparameters}.

\begin{figure}
    \centering
    \includegraphics[width=\columnwidth]{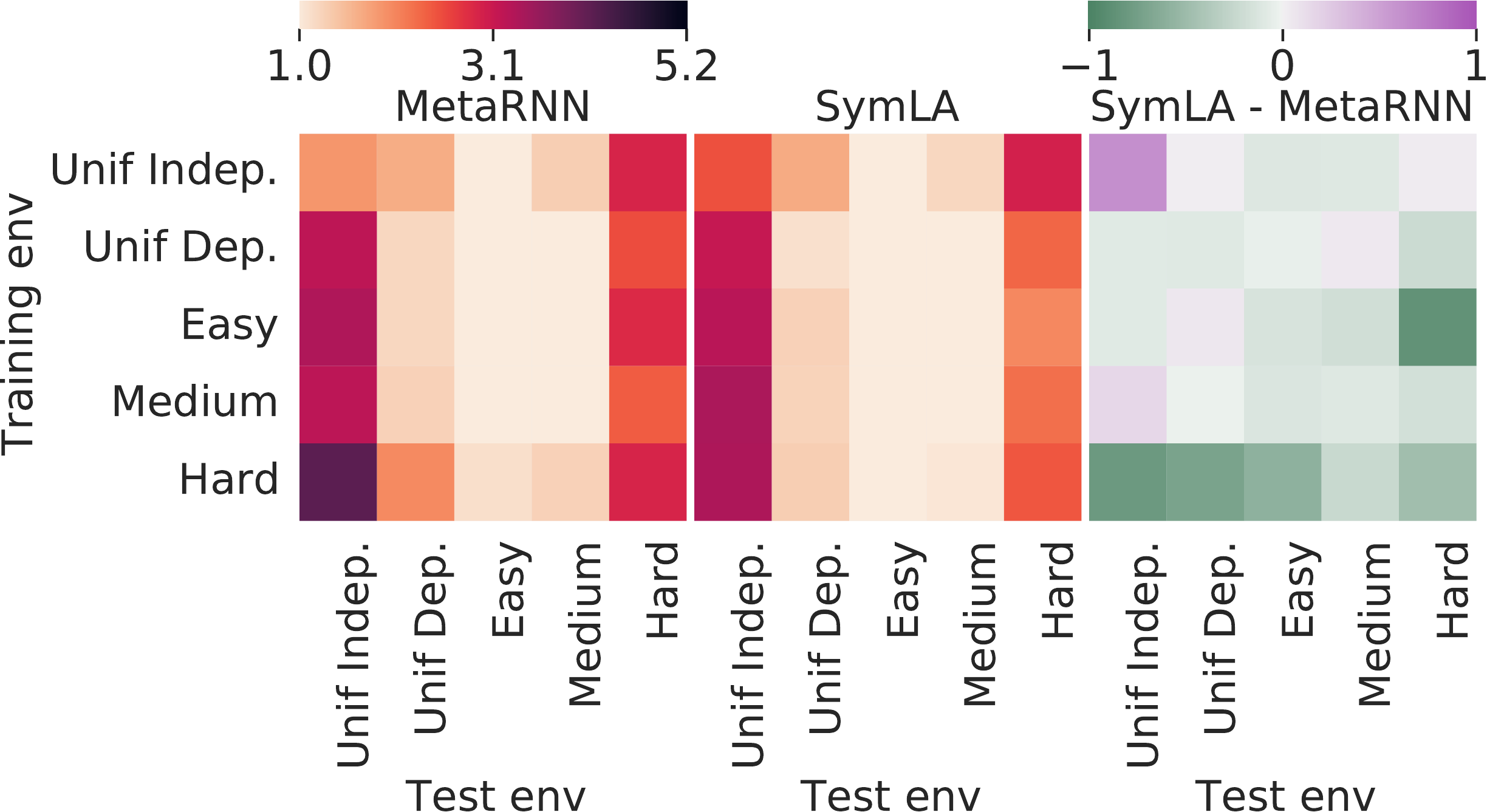}
    \caption{We compare SymLA to a standard MetaRNN on a set of bandit benchmarks from ~\citet{wang2016learning}.
    We train (y-axis) and test (x-axis) on two-armed bandits of varying difficulties.
    We report expected cumulative regret across 3 meta-training and 100 meta-testing runs with 100 arm-pulls (smaller is better).
    We observe that SymLA tends to perform comparably to the MetaRNN.
    }
    \label{fig:wang_bandits}
\end{figure}

\subsection{Learning to Learn on Similar Environments}

We first compare SymLA and the MetaRNN on the two-armed (dependent) bandit experiments from \citet{wang2016learning} where there is no large variation in the meta-test environments.
These consist of five different settings of varying difficulty that we use for meta-training and meta-testing (see Appendix \ref{app:wang_bandits}).
There are no observations (no context), only two arms, and a meta-training distribution where each arm has the same marginal distribution of payouts.
Thus, we expect the symmetries from SymLA to have no significant effect on performance.
We meta-train for an agent lifetime of $L = 100$ arm-pulls and report the expected cumulative regret at meta-test time in Figure \ref{fig:wang_bandits}.
We meta-train on each of the five settings, and meta-test across all settings.
The performance of the MetaRNN reproduces the average performance of \citet{wang2016learning}, here trained with ES instead of A2C.
When using symmetries (as in SymLA), we recover a similar performance compared to the MetaRNN.

\begin{figure}
    \centering
    \includegraphics[width=0.82\columnwidth]{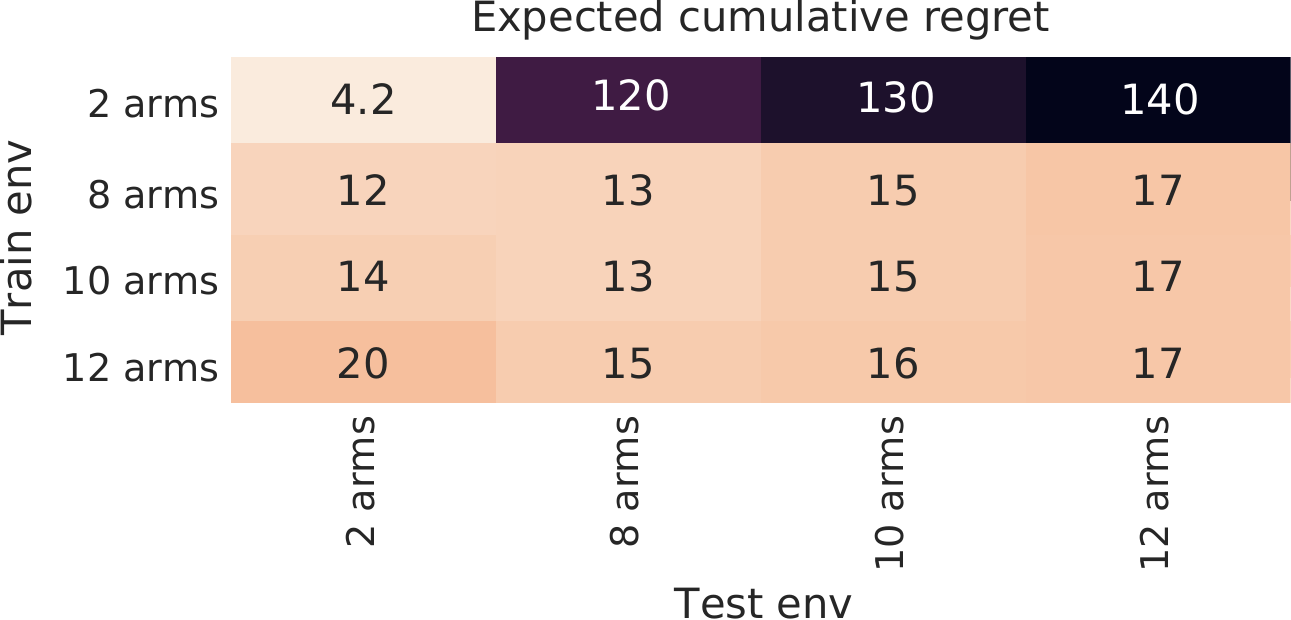}
    \caption{We meta-train and meta-test SymLA on varying numbers of independent arms to measure generalisation performance on unseen configurations.
    We do this by adding or removing RNNs to accommodate the additional output units.
    The number of meta-parameters remains constant.
    We report expected cumulative regret across 3 meta-training and 100 meta-testing runs with 100 arm-pulls (smaller is better). 
    Particularly relevant are the out-of-distribution scenarios (off-diagonal).
    }
    \label{fig:varying_arms_bandits}
\end{figure}

\begin{figure*}[h]
    \centering
    \includegraphics[width=0.8\textwidth]{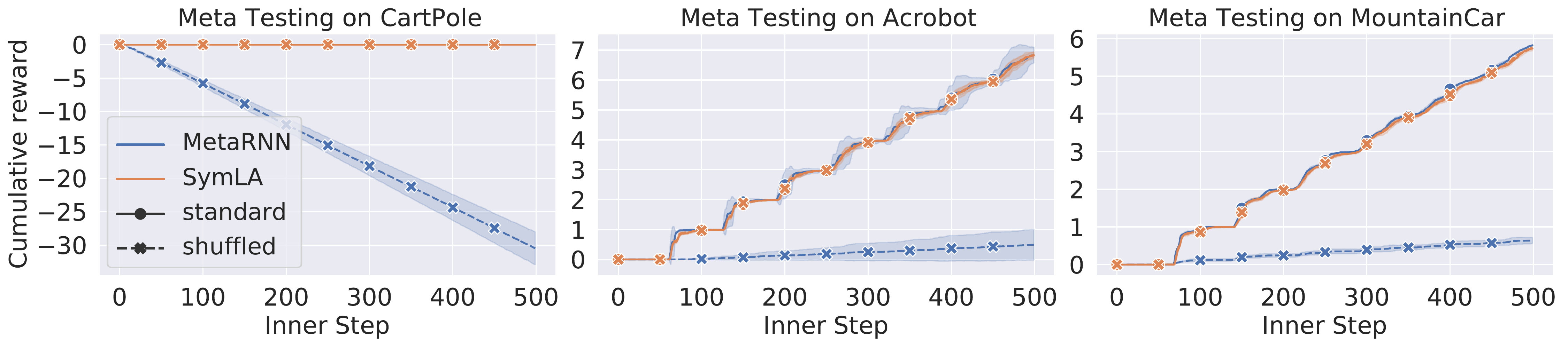}
    \caption{SymLA's architecture is inherently permutation invariant.
    When meta-training on standard CartPole, Acrobot, and MountainCar, the performance of the MetaRNN and SymLA are comparable.
    We then meta-test with shuffled observations and actions.
    In this setting, SymLA still performs well as it has meta-learned to identify observations and actions at meta-test time.
    In contrast, the MetaRNN fails to do so.
    Standard deviations are over 3 meta-training and 100 meta-testing runs.
    }
    \label{fig:classic_control_permute}
\end{figure*}

\begin{figure*}[h]
    \centering
    \raisebox{-0.44\height}{\includegraphics[width=0.3\textwidth]{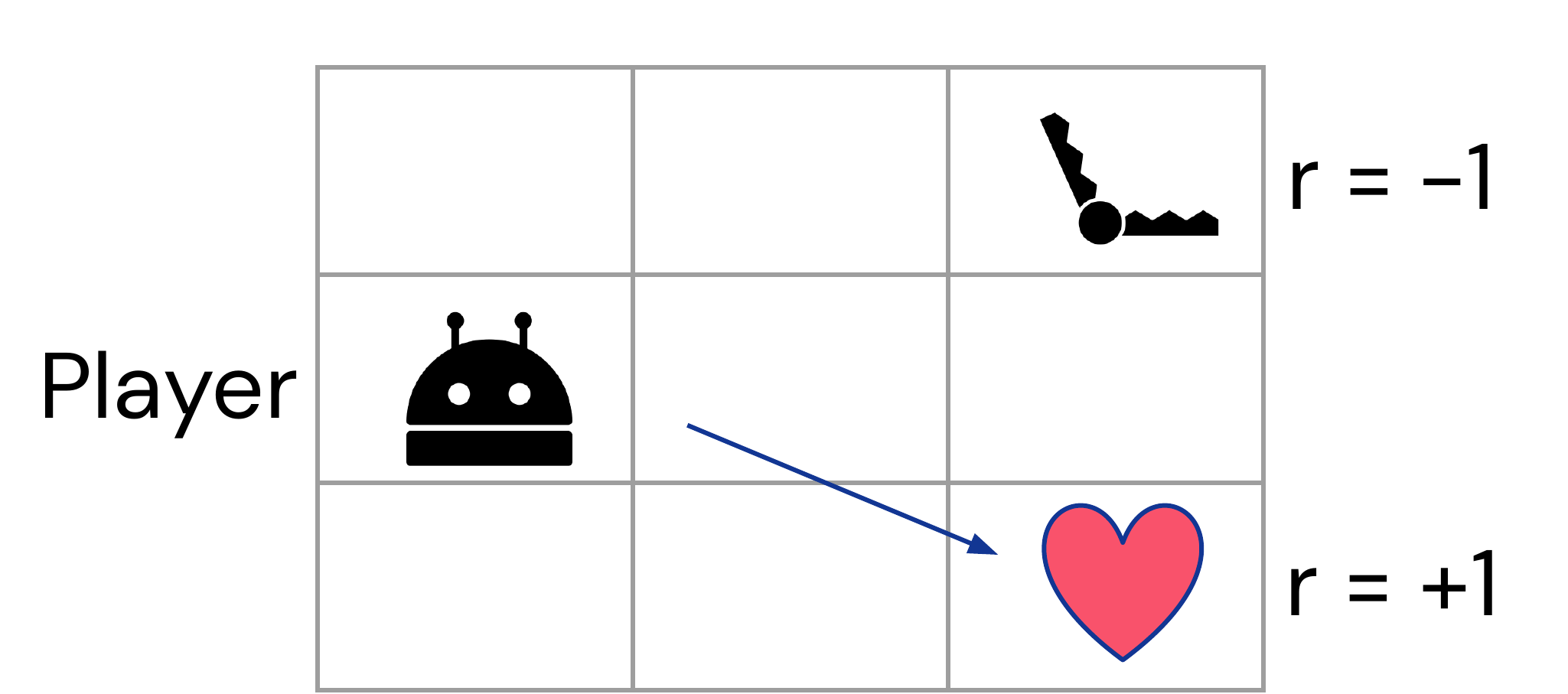}}
    \raisebox{-0.5\height}{\includegraphics[width=0.52\textwidth]{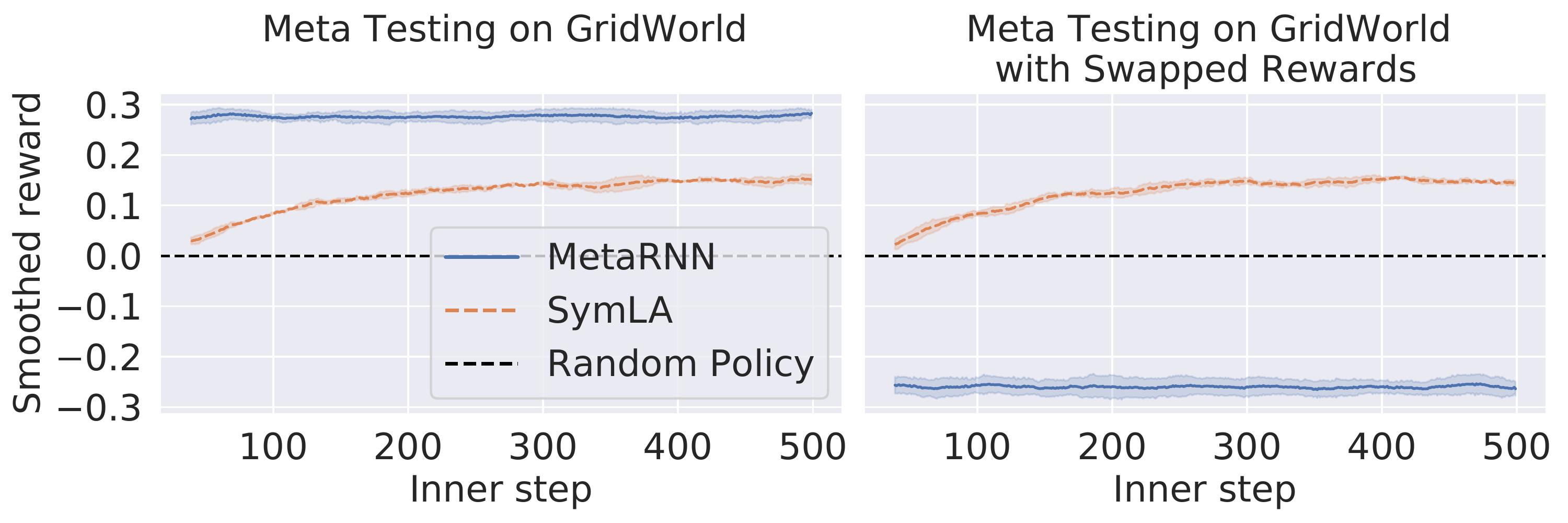}}
    \caption{
    We extend the permutation invariant property to concepts - varying the rewards associated with different object types (+1 and -1) in a grid world environment (left).
    SymLA is forced to learn about the rewards of object types at meta-test time (starting at near zero reward and increasing the reward intake over time).
    When switching the rewards and running the same learner, the MetaRNN collects the wrong rewards, whereas SymLA still infers the correct relationships.
    Standard deviations are over 3 meta-training and 100 meta-testing runs.
    }
    \label{fig:grid_concept_invariance}
\end{figure*}

\subsection{Generalisation to Unseen Action Spaces}

In contrast to the MetaRNN, in SymLA we can vary the number of arms at meta-test time.
The architecture of SymLA allows to change the network size arbitrarily by replicating existing RNNs, thus adding or removing arms at meta-test time while retaining the same meta-parameters from meta-training.
In Figure \ref{fig:varying_arms_bandits} we train on different numbers of arms and test on seen and unseen configurations.
All arms are independently drawn from the uniform distribution $p_i \sim U[0, 1]$.
We observe that SymLA works well within-distribution (diagonal)  and generalises to unseen numbers of arms (off-diagonal).
We also observe that for two arms a more specialized solution can be discovered, impeding generalisation when only training on this configuration.

\subsection{Generalisation to Unseen Observation Spaces}

In the next experiments we want to specifically analyze the permutation invariance created by our architecture.
In the previous bandit environments, actions occurred in all permutations in the training distribution.
In contrast, RL environments usually have some structure to their observations and actions.
For example in CartPole the first observation is usually the pole angle and the first action describes moving to the left.
Human-engineered learning algorithms are usually invariant to permutations and thus generalise to new problems with different structure.
The same should apply for our black-box agent with symmetries.
    
We demonstrate this property in the classic control tasks \emph{CartPole}, \emph{Acrobot}, and \emph{MountainCar}.
We meta-train on each environment respectively with the original observation and action order.
We then meta-test on either (1) the same configuration or (2) across a permuted version.
The results are visualized in Figure \ref{fig:classic_control_permute}.
Due to the built-in symmetries, the performance does not degrade in the shuffled setting.
Instead, our method quickly learns about the ordering of the relevant observations and actions at meta-test time.
In comparison, the MetaRNN baseline fails on the permuted setting where it was not trained on, indicating over-specialization.
Thus, symmetries help to generalise to observation permutations that were not encountered during meta training.

\subsection{Generalisation to Unseen Tasks}

The permutation invariance has further reaching consequences.
It extends to learning about tasks at meta-test time.
This enables generalisation to unseen tasks.
We construct a grid world environment (see Figure \ref{fig:grid_concept_invariance}) with two object types: A trap and a heart.
The agent and the two objects (one of each type) are randomly positioned every episode.
Collecting the heart gives a reward of +1, whereas the trap gives -1.
All other rewards are zero.
The agent observes its own position and the position of both objects.
The observation is constructed as an image with binary channels for the position and each object type.

When meta-training on this environment, at meta-test time we observe in Figure \ref{fig:grid_concept_invariance} that the MetaRNN learns to directly collect hearts in each episode throughout its lifetime. 
This is due to having overfitted to the association of hearts with positive rewards.
In comparison, SymLA starts with near-zero rewards and learns through interactions which actions need to be taken when receiving particular observations to collect the heart instead of the trap.
With sufficient environment interactions $L$ we would expect SymLA, if it implemented a general-purpose RL algorithm, to eventually (after sufficient learning) match the average reward per time of the MetaRNN in the non-shuffled grid world.
Next, we swap the rewards of the trap and heart, i.e. the trap now gives a positive reward, whereas the heart gives a negative reward.
This is equivalent to swapping the input channels corresponding to the heart and trap.
We observe that SymLA still generalises, learning at meta-test time about observations and their associated rewards.
In contrast, the MetaRNN now collects the wrong item, receiving negative rewards.
These results show that black-box meta RL with symmetries discovers a more general update rule that is less specific to the training tasks than typical MetaRNNs.

\begin{figure}
    \centering
    \includegraphics[width=0.47\textwidth]{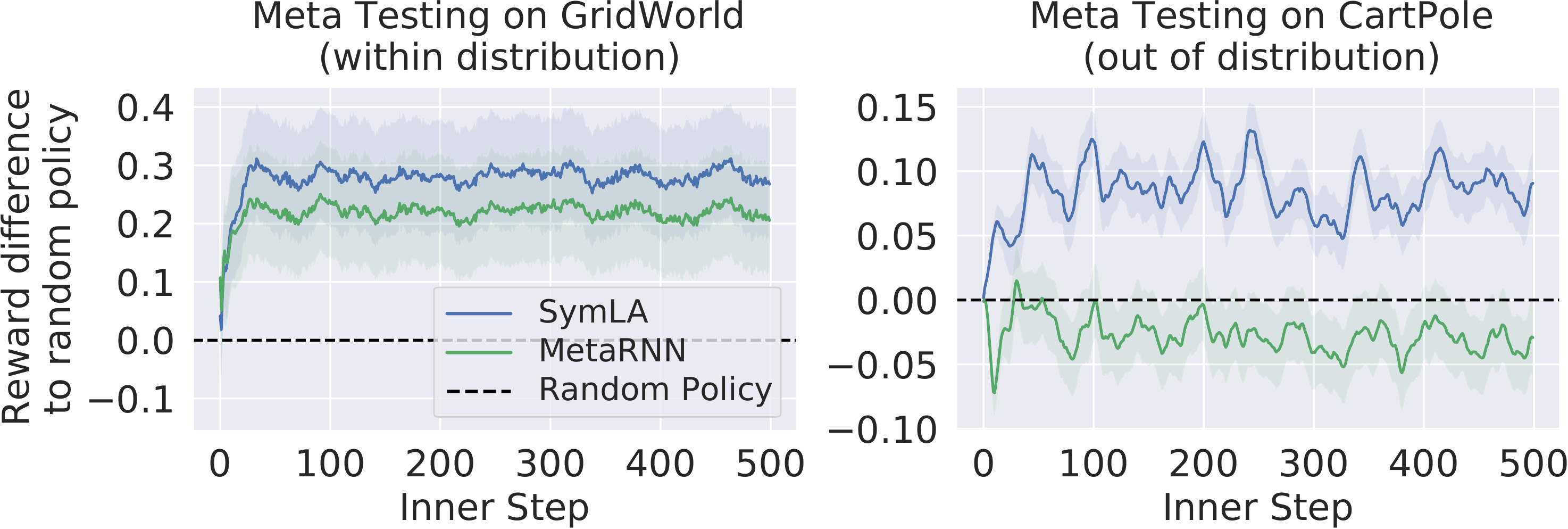}
    \caption{
    Generalisation capabilities of SymLA from GridWorld to CartPole.
    We meta-train the learning algorithm on GridWorld.
    We then meta-test on GridWorld and CartPole and report standard error of the mean and mean rewards (100 seeds) relative to a random policy - this highlights the learning process.
    While SymLA generalises from GridWorld to CartPole, the MetaRNN does not.
    }
    \label{fig:unseen_environments}
\end{figure}

\subsection{Generalisation to Unseen Environments}
We have demonstrated how permutation invariance can lead to increased generalisation.
But can SymLA also generalise between entirely different environments?
We show-case how meta-training on a grid world environment allows generalisation to CartPole.
To simplify credit-assignment, we use a dense-reward grid world where the reward is proportional to the change in distance toward a target position.
Both the target position, as well as the agent position are randomized.
The agent observes its own position, all obstacles, and the target position as a binary image with multiple channels.
In the CartPole environment the agent is rewarded for being as upright and centered as possible~\citep{tassa2020dmcontrol}.
Further, during meta-training, we randomly project observations linearly for each lifetime.
This is necessary as in the grid world environment all observations are binary whereas the CartPole environment has continuously varying observations.
This mismatch would inhibit generalisation.
In Figure \ref{fig:unseen_environments} we demonstrate that meta-training with SymLA only on the GridWorld environment allows reusing the same meta-learned learning algorithm to the CartPole environment.
In contrast, the MetaRNN does not exhibit such generalisation.
This suggests that meta learning with symmetries has the potential to produce learning algorithms that generalize between significantly different environments.
    
\section{Related Work}

\paragraph{Black-Box Meta RL}
Black-box meta RL can be implemented by policies that receive the reward signal as input~\citep{schmidhuber1993self} and use memory to learn, such as recurrence in RNNs~\citep{hochreiter2001learning,wang2016learning,duan2016rl}.
These approaches do not feature the symmetries discussed in this paper which leads to a tendency of overfitting.

\paragraph{Learned Learning Rules \& Fast Weights}
In the supervised and reinforcement learning contexts, learned learning rules~\citep{bengio1992optimization} or fast weights~\citep{schmidhuber1992learning,schmidhuber1993reducing,miconi2018differentiable,schlag2020learning,najarro2020meta} describe (meta-)learned mechanisms (slow weights) that update fast weights to implement learning.
This often involves outer-products and can be generalised to black-box meta learning with parameter sharing~\citep{kirsch2020meta}.
None of these approaches feature all of the symmetries we discuss above to meta learn RL algorithms.

\paragraph{Backpropagation-based Meta RL}
Alternatives to black-box meta RL include learning a weight initialization and adapting it with a human-engineered RL algorithm~\citep{finn2017model}, warping computed gradients~\citep{flennerhag2019meta}, meta-learning hyper-parameters~\citep{sutton1992adapting,Xu:2018} or meta-learning objective functions corresponding to the learning algorithm~\citep{houthooft2018evolved,kirsch2019improving,Xu:2020,oh2020discovering,bechtle2021meta}.

\paragraph{Neural Network Symmetries}
Symmetries in neural networks have mainly been investigated to reflect the structure of the input data.
This includes applications of convolutions~\citep{fukushima1979neural}, deep sets~\citep{zaheer2017deep}, graph neural networks~\citep{wu2020comprehensive}, geometric deep learning~\citep{bronstein2017geometric}, or meta learning symmetries~\citep{zhou2021metalearning}.
In contrast, our work focuses on the structure and symmetries of learning algorithms.
While many meta learning algorithms exhibit symmetries~\citep{bengio1992optimization}, in particular backpropagation-based meta learning~\citep{andrychowicz2016learning,finn2017model,flennerhag2019meta,kirsch2019improving}, the effects of these symmetries have not been discussed in detail.
In this work, we provide such a discussion and experimental investigation in the context of meta RL.

\section{Conclusion}

In this work, we identified symmetries that exist in backpropagation-based methods for meta RL but are missing from black-box methods.
We hypothesized that these symmetries lead to better generalisation of the resulting learning algorithms.
To test this, we extended a black-box meta learning method~\citep{kirsch2020meta} that exhibits these same symmetries to the meta RL setting. 
This resulted in SymLA, a flexible black-box meta RL algorithm that is less prone to over-fitting compared to MetaRNNs.
We demonstrated generalisation to varying numbers of arms in bandit experiments (unseen action spaces), permuted observations and actions with no degradation in performance (unseen observation spaces), and observed the tendency of the meta-learned RL algorithm to learn about states and their associated rewards at meta-test time (unseen tasks).
Finally, we showed that the discovered learning behavior also transfers between grid world and (unseen) classic control environments.

\section*{Acknowledgements}

We thank Nando de Freitas, Razvan Pascanu, and Luisa Zintgraf for helpful comments.
Funded by DeepMind.

\bibliography{main}

\clearpage

\appendix

\begin{table*}[]
\small
\caption{
    A comparison between fixed reinforcement learning algorithms (REINFORCE), backpropagation-based meta RL (MAML, MetaGenRL, LPG), black-box (MetaRNN), and our black-box method with symmetries (SymLA).
    $\pi_\theta^{(s)}$ denotes a \emph{stationary} policy that is updated at fixed intervals by backpropagation.
}
\begin{tabular}{@{}c|ccccc@{}}
\toprule
                                 & REINFORCE                                                                & MetaGenRL / LPG                                                & MAML                                                                     & MetaRNN       & SymLA (ours)                    \\ \midrule
Meta variables             & /                                                                        & $\phi$                                                                  & Initial $\theta_0$                                                       & $\theta$      & $\theta$                  \\
Learned variables           & $\theta$                                                                 & $\theta$                                                                & $\theta$                                                                 & RNN state $h$ & RNN states $h_{ab}^{(k)}$ \\
Learning algorithm               & \begin{tabular}[c]{@{}c@{}}fixed loss func $L$\\ + Backprop\end{tabular} & \begin{tabular}[c]{@{}c@{}}learned loss func $L_\phi$\\ + Backprop\end{tabular} & \begin{tabular}[c]{@{}c@{}}fixed loss func $L$\\ + Backprop\end{tabular} & $\pi_\theta$  & $\pi_\theta$              \\
Policy                           & $\pi_\theta^{(s)}$                                                       & $\pi_\theta^{(s)}$                                                      & $\pi_\theta^{(s)}$                                                             & $\pi_\theta$  & $\pi_\theta$              \\
Black-box                        & \no                                                                      & \no                                                                     & \no                                                                      & \yes          & \yes                      \\
Symmetries in learning algorithm & \yes                                                                     & \yes                                                                    & \yes                                                                   & \no           & \yes                      \\ \bottomrule
\end{tabular}
\end{table*}

\begin{algorithm*}
    \centering	
    \begin{algorithmic}	
        \Require Distribution over RL environment(s) $p(e)$
        \State $\theta \leftarrow$ initialize LSTM parameters
        \While{meta loss has not converged} \Comment{Outer loop in parallel over envs $e \sim p(e)$ and samples $\phi \sim \N(\phi|\theta, \Sigma)$}
            \State $\{h_{ab}\} \leftarrow$ initialize LSTM states $\quad \forall a,b$
            \State $o_1 \sim p(o_1)$ \Comment{Initialize environment $e$}
            \For{$t \in \{1, \ldots, L\}$} \Comment{Inner loop over lifetime in environment $e$}
                \State $h_{ab} \leftarrow f_{\textrm{LSTM}}(h_{ab}, o_{t,a}, a_{t-1,b}, r_{t-1}, \fmsg_b, \bmsg_a) \quad \forall a,b$ \Comment{Equation \ref{eq:state_update}}
                \State $\fmsg_b \leftarrow \sum_a f_{\fmsg}(h_{ab}) \quad \forall b$ \Comment{Create forward messages}
                \State $\bmsg_a \leftarrow \sum_b f_{\bmsg}(h_{ab}) \quad \forall a$ \Comment{Create backward messages}
                \State $y \leftarrow \fmsg_{\cdot1}$ \Comment{Read out action}
                \State $a_t \sim p(a_t; y)$ \Comment{Sample action from distribution parameterized by $y$}
                \State Send action $a_t$ to environment $e$, observe $o_{t+1}$ and $r_t$
            \EndFor
            \State $\theta \leftarrow \theta + \alpha \nabla_\theta \E_{\phi \sim \mathcal{N}(\phi|\theta, \Sigma)}[\E_{e \sim p(e)}[\sum_{t=1}^L r_t^{(e)}(\phi)]]$ \Comment{Update $\theta$ using evolution strategies (Equation \ref{eq:evolution_strategies})}
        \EndWhile	
    \end{algorithmic}	
    \caption{SymLA meta training}\label{alg:meta_training}
\end{algorithm*}

\section{Bandits from \citet{wang2016learning}}\label{app:wang_bandits}
In our experiments, we use bandits of varying difficulty from \citet{wang2016learning}.
Let $p_1$ be the probability of the first arm for a payout of $r = 1$, $r = 0$ otherwise, and $p_2$ the payout for the second arm.
Then, we define the
\begin{itemize}
    \item uniform independent bandit with $p_1 \sim U[0,1]$ and $p_2 \sim U[0,1]$,
    \item uniform dependent bandit with $p_1 \sim U[0,1]$ and $p_2 = 1 - p_1$,
    \item easy dependent bandit with $p_1 \sim U\{0.1, 0.9\}$ and $p_2 = 1 - p_1$,
    \item medium dependent bandit with $p_1 \sim U\{0.25, 0.75\}$ and $p_2 = 1 - p_1$,
    \item hard dependent bandit with $p_1 \sim U\{0.4, 0.6\}$ and $p_2 = 1 - p_1$.
\end{itemize}

\section{Hyper-parameters}\label{app:hyperparameters}

\subsection{SymLA Architecture}
We use a single recurrent layer, $K = 1$, with a message size of $\overleftarrow M = 8$ and $\overrightarrow M = 8$.
To produce the next state $h_{ab}$ according to Equation \ref{eq:state_update}, we use parameter-shared LSTMs with a hidden size of $N = 16$ ($N = 64$ for bandits to match \citet{wang2016learning}) and run the recurrent cell for 2 micro ticks.

\subsection{Meta Learning / Outer Loop}
We estimate gradients $\nabla_\theta$ using evolutionary strategies~\citep{salimans2017evolution} with $10$ evaluations per population sample to estimate the fitness value ($100$ evaluations for bandits).
Then, we apply those using Adam with a learning rate of $\alpha = 0.01$, $\beta_1 = 0.9$, and $\beta_2 = 0.999$ ($\alpha = 0.2$ for bandits).
We use a fixed noise standard deviation of $\sigma = 0.035$ ($\sigma = 0.2$ for bandits) and a population size of $512$.
Our inner loop has a length of $L=500$ ($L=100$ for bandits), concatenating multiple episodes.
We meta-optimize for $4,000$ outer steps for bandit experiments, and $20,000$ otherwise.

\subsection{Generalisation to Unseen Environments}

We apply a random linear transformation (Glorot normal) to environment observations, mapping those to a $16$-dimensional vector.

\section{Scalability and complexity}
The computational complexity of the inner loop (and meta testing) is $O(N^2W)$ per environment step, where $N$ is the hidden size of each RNN and $W$ is the number of RNNs (number of parameters in a conventional neural network).
$N$ can generally be small, in most experiments $N = 16$ (see Appendix \ref{app:hyperparameters}).
Memory complexity is independent of the number of time-steps and is $O(N^2+NW+MS)$, where $M = 8$ is the message size and $S$ denotes the number of messages.
The computational complexity of meta training highly depends on the chosen meta-optimizer.
With ES, each outer optimization step has a complexity of $O(N^2WLPE)$, where $L = 500$ is the length of the evaluated lifetime, $P = 512$ is the size of the particle population (within range of the ES literature), and E is the number of evaluations per particle to estimate the average reward.
Memory complexity is generally low for gradient-free optimization such as ES; for meta-training it is $O(N^2+NW+MS)$, if the population is evaluated in sequence.
Compared to the MetaRNN (RL$^2$), SymLA is slower by a factor of $N^2$ (here $N=16$) in both meta training and testing if the number of RNNs is chosen to equal the MetaRNN's parameters.
In practice, the RNNs also increase the capacity such that fewer RNNs may also be sufficient.

\newpage
\section{Code snippet}
\definecolor{dkgreen}{rgb}{0,0.6,0}
\definecolor{gray}{rgb}{0.5,0.5,0.5}
\definecolor{mauve}{rgb}{0.58,0,0.82}
\lstset{
  language=Python,
  aboveskip=3mm,
  belowskip=3mm,
  showstringspaces=false,
  keepspaces=true,
  columns=flexible,
  basicstyle={\ttfamily\tiny},
  numbers=none,
  numberstyle=\tiny\color{gray},
  keywordstyle=\color{blue},
  commentstyle=\color{dkgreen},
  stringstyle=\color{mauve},
  breaklines=true,
  breakatwhitespace=true,
  tabsize=3
}
\begin{lstlisting}[language=Python]
import haiku as hk
import jax
import jax.numpy as jnp
import numpy as np
import optax


class SymlaLayer(hk.Module):

  def __init__(self, input_size: int, output_size: int, msg_size: int, hidden_size: int, micro_ticks: int):
    super().__init__()
    self.input_size = input_size
    self.output_size = output_size
    self.micro_ticks = micro_ticks
    self._lstm = hk.LSTM(hidden_size)
    self._fwd_messenger = hk.Linear(msg_size)
    self._bwd_messenger = hk.Linear(msg_size)
    self._tick = hk.vmap(hk.vmap(self._tick, (0, None, 0, None)), (0, 0, None, None))

  def _tick(self, lstm_state: hk.LSTMState, fwd_msg: jnp.ndarray, bwd_msg: jnp.ndarray, aux: jnp.ndarray):
    inp = jnp.concatenate([fwd_msg, bwd_msg, aux])
    out, lstm_state = self._lstm(inp, lstm_state)
    return out, lstm_state

  def create_state(self):
    lstm_state_shape = (2, self.input_size, self.output_size, self._lstm.hidden_size)
    lstm_state = jnp.zeros(lstm_state_shape)
    lstm_state = hk.LSTMState(hidden=lstm_state[0], cell=lstm_state[1])

    fwd_msg_shape = (self.output_size, self._fwd_messenger.output_size)
    fwd_msg = jnp.zeros(fwd_msg_shape)

    bwd_msg_shape = (self.input_size, self._bwd_messenger.output_size)
    bwd_msg = jnp.zeros(bwd_msg_shape)

    return lstm_state, fwd_msg, bwd_msg

  def __call__(self, state, inp: jnp.ndarray, inp_end: jnp.ndarray, aux: jnp.ndarray):
    lstm_state, fwd_msg, bwd_msg = state

    # Update state
    in_fwd_msg = jnp.concatenate([bwd_msg, inp[:, None]], axis=-1)
    in_bwd_msg = jnp.concatenate([fwd_msg, inp_end[:, None]], axis=-1)
    for _ in range(self.micro_ticks):
      out, lstm_state = self._tick(lstm_state, in_fwd_msg, in_bwd_msg, aux)

    # Update forward messages
    out_fwd_msg = self._fwd_messenger(out).mean(axis=0)
    # Update backward messages
    out_bwd_msg = self._bwd_messenger(out).mean(axis=1)

    # Read out logits for action
    logits = out_fwd_msg[:, 0]

    return logits, (lstm_state, out_fwd_msg, out_bwd_msg)


class SymlaModel(hk.Module):

  def __init__(self):
    super().__init__()
    self._layer = SymlaLayer(...)

  def __call__(self, env, env_state):
    prev_action = jnp.zeros(env.action_shape)
    state = self._layer.create_state()
    rng_ticks = jnp.array(hk.next_rng_keys(env.meta_episode_length))

    def scan_tick(carry, rng_tick):
      env_state, state, prev_action = carry
      rng_tick, rng_action = jax.random.split(rng_tick)

      # Obtain signals from environment
      obs = env.observation(env_state)
      reward = env.reward(env_state)
      done = env.is_terminal(env_state).astype(jnp.float32)

      # Tick layer
      inp = obs.flatten()
      aux = jnp.stack([reward, done])
      logits, new_state = self._layer(state, inp, prev_action, aux)

      # Create action
      action = jax.random.categorical(rng_action, logits)
      action = jax.nn.one_hot(action, logits.shape[-1])

      # Tick environment
      new_env_state = env.step(rng_tick, env_state, action)
      reward = env.reward(env_state)

      return (new_env_state, new_state, action), reward

    _, rewards = hk.scan(scan_tick, (env_state, state, prev_action), rng_ticks)
    loss = -jnp.mean(rewards)
    return loss, rewards


class Experiment:

  def __init__(self, noise_std: float, population_size: int, learning_rate: float):
    self._model = hk.transform(lambda *x: SymlaModel()(*x))
    self._optimizer = optax.adam(learning_rate)
    self._env = Env()
    self._population_size = population_size
    self._noise_std = noise_std
    self._update_func = jax.jit(self._update_func)

  def _es_eval(self, params, rng, env_state):
    # Extract shapes
    treedef = jax.tree_structure(params)
    shapes = jax.tree_map(lambda p: np.asarray(p.shape), params)

    # Random keys
    rng, param_rng = jax.random.split(rng)
    keys = jax.tree_unflatten(treedef, jax.random.split(param_rng, treedef.num_leaves))

    # Generate noise
    noise = jax.tree_multimap(jax.random.normal, keys, shapes)
    scaled_noise = jax.tree_map(lambda x: x * self._noise_std, noise)

    # Antithetic sampling
    params_pos = jax.tree_multimap(jnp.add, params, scaled_noise)
    params_neg = jax.tree_multimap(jnp.subtract, params, scaled_noise)

    # Evalute in environment
    loss_pos, rewards = self._model.apply(params_pos, rng, self._env, env_state)
    loss_neg, _ = self._model.apply(params_neg, rng, self._env, env_state)

    # Compute grads
    es_factor = (loss_pos - loss_neg) / (2 * self._noise_std ** 2)
    grads = jax.tree_map(lambda x: x * es_factor, scaled_noise)

    return grads

  def _update_func(self, params, opt_state, rng):
    rng, rng_update = jax.random.split(rng)
    grads = self._es_grads(params, rng_update)

    updates, opt_state = self._optimizer.update(grads, opt_state)
    params = optax.apply_updates(params, updates)

    return params, opt_state, rng

  def _es_grads(self, params, rng):
    rng_env_init, rng_eval = jax.random.split(rng)
    rng_env_init = jax.random.split(rng_env_init, self._population_size)
    rng_eval = jax.random.split(rng_eval, self._population_size)

    env_state = jax.vmap(self._env.initial_state)(rng_env_init)
    v_es_eval = jax.vmap(self._es_eval, in_axes=(None, 0, 0))
    grads = v_es_eval(params, rng_eval, env_state)
    grads = jax.tree_map(lambda x: jnp.mean(x, axis=0), grads)

    return grads

  def train(self, seed: int, num_iterations: int):
    rng = jax.random.PRNGKey(seed)
    rng, rng_init = jax.random.split(rng)

    dummy_env_state = self._env.initial_state(rng_init)

    params = self._model.init(rng_init, self._env, dummy_env_state)
    opt_state = self._optimizer.init(params)

    for _ in range(num_iterations):
      params, opt_state, rng = self._update_func(params, opt_state, rng)
\end{lstlisting}

\end{document}